\documentclass[letterpaper, 10 pt, conference]{ieeeconf} 
\IEEEoverridecommandlockouts                                  
\usepackage{times}
\usepackage{epsfig}
\usepackage{graphicx}
\usepackage{amsmath}
\usepackage{amssymb}
\usepackage{multirow}
\usepackage{comment}
\usepackage{color}
\usepackage{multirow}
\usepackage{makecell}
\usepackage{times}
\usepackage{bbding}
\usepackage{pifont}
\usepackage{float}
\usepackage{cite}
\usepackage{threeparttable}

\usepackage[hidelinks,colorlinks=true,linkcolor=blue,citecolor=blue,urlcolor=blue]{hyperref}
\usepackage{booktabs}
\usepackage[linesnumbered,ruled,vlined]{algorithm2e} 
\SetKwInput{KwIn}{Input}    
\SetKwInput{KwOut}{Output}

\title{\LARGE \bf
AirVista-II: An Agentic System for Embodied UAVs Toward \\Dynamic Scene Semantic
Understanding}

\author{
Fei Lin$^{1}$ $^{*}$, Yonglin Tian$^{2}$ $^{*}$, Tengchao Zhang$^{1}$, Jun Huang$^{1}$, Sangtian Guan$^{1}$, and Fei-Yue Wang$^{2, 1}$ $^{\dagger}$ 
\thanks{$^{*}$Fei Lin and Yonglin Tian contribute equally to this work.}
\thanks{$^{\dagger}$Corresponding author: Fei-Yue Wang (email: feiyue.wang@ia.ac.cn).}
\thanks{This work was partly supported by the Science and Technology Development Fund, Macau Special Administrative Region (SAR) (0145/2023/RIA3, 0093/2023/RIA2, 0157/2024/RIA2)}
\thanks{$^{1}$Fei Lin, Tengchao Zhang, Jun Huang, and Sangtian Guan are with the Department of Engineering Science, Faculty of Innovation Engineering, Macau University of Science and Technology, Macau 999078, China (e-mail: \{feilin, zhangtengchao, junhuang, sangtian.guan\}@ieee.org).}
\thanks{$^{2}$Yonglin Tian is with the State Key Laboratory for Management and Control of Complex Systems, Institute of Automation, Chinese Academy of Sciences, Beijing 100190, China (e-mail: yonglin.tian@ia.ac.cn).}
\thanks{$^{2, 1}$Fei-Yue Wang is with the State Key Laboratory for Management and Control of Complex Systems, Chinese Academy of Sciences, Beijing 100190, and also with the Department of Engineering Science, Faculty of Innovation Engineering, Macau University of Science and Technology, Macau 999078, China (e-mail: feiyue.wang@ia.ac.cn).}%
}

\begin{document}

\maketitle
\thispagestyle{empty}
\pagestyle{empty}


\begin{abstract}
Unmanned Aerial Vehicles (UAVs) are increasingly important in dynamic environments such as logistics transportation and disaster response. However, current tasks often rely on human operators to monitor aerial videos and make operational decisions. This mode of human-machine collaboration suffers from significant limitations in efficiency and adaptability. In this paper, we present AirVista-II—an end-to-end agentic system for embodied UAVs, designed to enable general-purpose semantic understanding and reasoning in dynamic scenes. The system integrates agent-based task identification and scheduling, multimodal perception mechanisms, and differentiated keyframe extraction strategies tailored for various temporal scenarios, enabling the efficient capture of critical scene information. Experimental results demonstrate that the proposed system achieves high-quality semantic understanding across diverse UAV-based dynamic scenarios under a zero-shot setting.
\end{abstract}


\section{Introduction}

In dynamic environments, UAVs typically rely on onboard cameras and other sensors to acquire real-time images and video streams in support of task-level decision-making. Semantic understanding tasks require UAVs to function not only as data acquisition platforms, but also to possess embodied intelligence—enabling semantic modeling of the environment and natural language interaction—so as to generate high-level semantic responses that are responsive to human operational instructions based on perceived information~\cite{wu2024embodied}.

In recent years, Foundation Models (FMs), represented by Large Language Models (LLMs), have demonstrated strong autonomy and domain adaptability in the field of embodied intelligence~\cite {tian2025algorithmic, zheng2024computational, tian2024logisticsvista}. Numerous studies have explored the use of LLMs to enhance UAV understanding in dynamic scenarios. For example, de Zarzà \emph{et al}.~\cite{de2023socratic}, and de Curtò \emph{et al}.~\cite{de2023semantic} integrated multiple FMs to construct UAV video understanding systems. These approaches have achieved significant progress in semantic modeling; however, they generally lack explicit task-planning mechanisms, leading to instability in response controllability. Moreover, due to the absence of external tool invocation capabilities and a coordinated multi-module framework, their generalization remains limited when dealing with structurally complex and open-ended tasks~\cite{zhao2025cityeqa}.

To address the above challenges, the recently emerging agentic Artificial Intelligence (AI) offers a more flexible and general system paradigm. This paradigm is driven by LLMs and possesses core capabilities such as task decomposition, tool invocation, and multi-agent coordination and scheduling~\cite{shavit2023practices}. It holds promise for offering a unified framework supporting UAV-performed semantic understanding tasks in dynamic scenes~\cite{tian2025uavs}. In this paper, we construct an agentic system for embodied UAVs, which uniformly categorizes dynamic scenes into three task forms: images, short videos, and long videos. For each task form, we design corresponding agent workflows and model invocation strategies to achieve a complete autonomous process covering task recognition, data parsing, semantic understanding, and high-level reasoning.

The main contributions of this work are as follows:
\begin{itemize}
    \item We propose AirVista-II, an end-to-end agentic system for embodied UAVs, promoting the transition of UAVs from passive data acquisition to an active paradigm of semantic interaction.
    \item We design an adaptive keyframe extraction strategy for long-video scenarios, which effectively captures semantically salient frames and enhances the UAV’s ability to understand complex dynamic scenes.
    \item We validate the proposed approach on multiple public aerial video datasets, demonstrating high accuracy and descriptive quality under zero-shot settings.
\end{itemize}

\section{Preliminaries}

\subsection{Video Language Models}

Video Language Models (VideoLMs) extend Vision-Language Models (VLMs) by incorporating temporal sequence modeling to enhance the understanding of dynamic information in videos. Existing implementations of VideoLMs can be broadly categorized into two approaches: The first integrates video and language modalities directly during training, with representative methods including Video-LLaVA~\cite{lin2023video} and Video-ChatGPT~\cite{maaz2023video}, among others; the second leverages existing VLM architectures by transforming videos into structured image inputs through multi-stage processing. For example, IG-VLM~\cite{kim2024image} samples 6 frames at equal intervals from a video and arranges them into a grid-like image, which is then fed into a standard VLM to model temporal information. Similar approaches include VideoTree~\cite{wang2024videotree}, LVNet~\cite{park2024too}, and others.

\subsection{Embodied UAVs for Dynamic Scene Understanding}

Existing studies can be broadly categorized into two representative paradigms. The first approach directly integrates UAVs with one or more FMs to enable basic perception and question-answering capabilities, with representative works including~\cite{de2023socratic, de2023semantic}, among others. The second approach incorporates FMs as agents within UAV systems, forming a more comprehensive perception–understanding–decision loop. For example, the PMA agent proposed by Zhao \emph{et al}.~\cite{zhao2025cityeqa} is driven by a VLM and integrates three modules—Planner, Manager, and Actor—enabling UAVs to achieve strong understanding capabilities in dynamic scenes. Similar efforts include AeroVerse~\cite{yao2024aeroverse}, Patrol Agent~\cite{yuan2024patrol}, and our previously proposed AirVista~\cite{lin2024airvista}, among others.

Although the above methods share similar objectives with the approach proposed in this paper, they differ fundamentally in task paradigms and system architecture. First, existing works primarily focus on static images or short-term memory enhancement. In contrast, our core task addresses dynamic scene understanding across multiple temporal scales—including images, short videos, and long videos. Secondly, in terms of system architecture, we integrate multiple LLM-based agents into an agentic system equipped with both task planning and execution capabilities. This system serves as the ``brain'' of the UAV embodied agent, enabling multi-stage automated decision-making and execution for complex tasks. The design is inspired by the Agentic UAV~\cite{tian2025uavs} framework proposed by Tian \emph{et al}., and this work can be regarded as a concrete task-level implementation of that framework.

\subsection{UAV-based Dynamic Scene Datasets}

In this work, we focus on three UAV-based dynamic scene datasets to construct our experimental scenarios. Event Recognition in Aerial Videos (ERA) \cite{mou2020era} contains 2,864 aerial videos captured by UAVs. Each clip is approximately 5 seconds long and serves as a benchmark for video event recognition from a UAV perspective. CapERA \cite{bashmal2023capera} can be regarded as an extended version of ERA, where five human-annotated captions are provided for each aerial video, supporting tasks such as automatic video description and subtitle generation for UAV footage. SynDrone \cite{rizzoli2023syndrone} is generated in a simulated environment and includes aerial data collected from eight urban scenes (Town01–Town07 and the highly complex Town10HD) at three different altitudes: 20 meters, 50 meters, and 80 meters.


\section{Approach}

\begin{figure*}
    \centering
    \includegraphics[width=\textwidth]{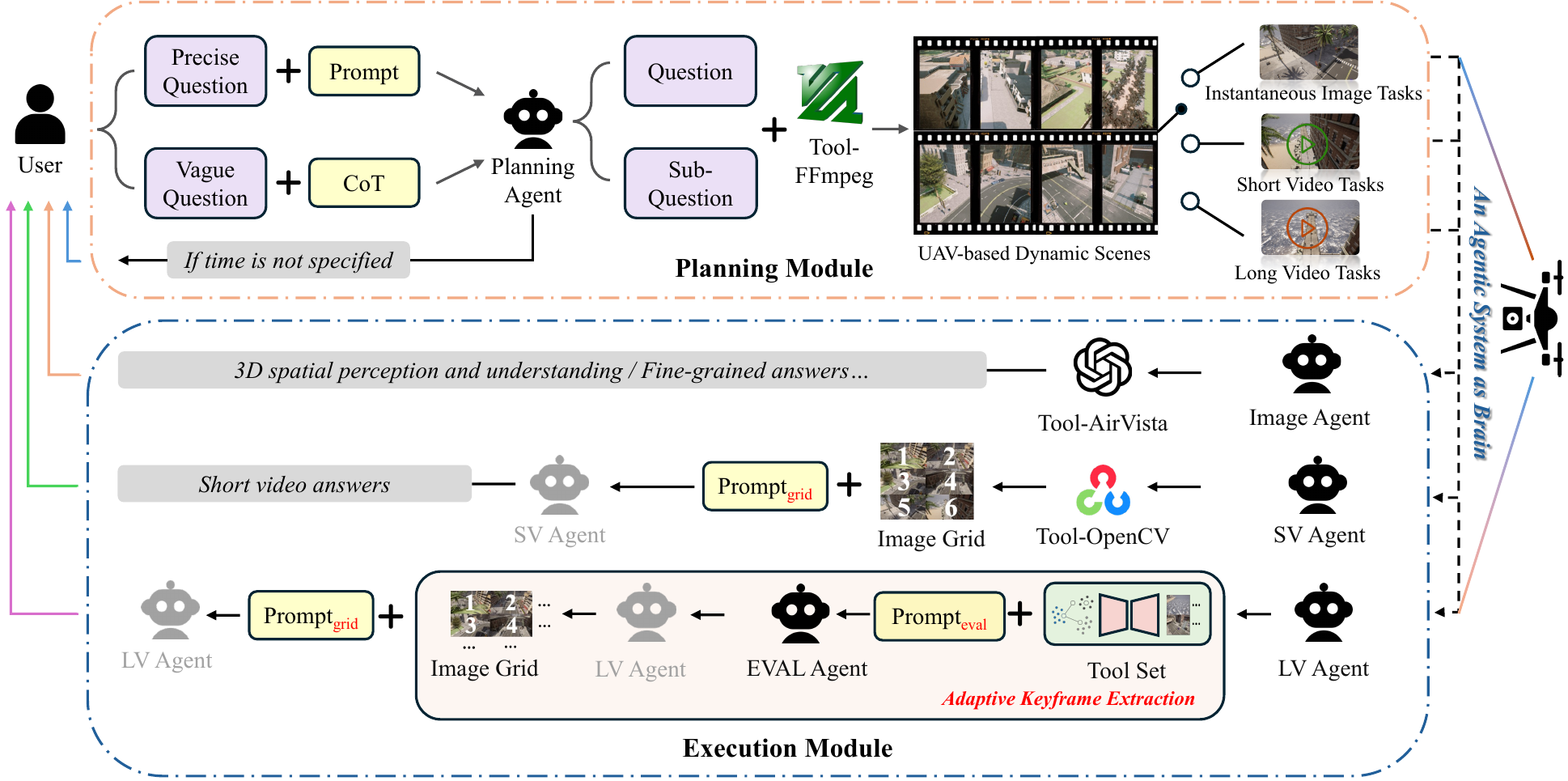}
    \caption{Execution pipeline of the AirVista-II system. For clarity, \textit{Long Video Agent}, \textit{Short Video Agent}, and \textit{Evaluation Agent} are abbreviated as \textbf{LV Agent}, \textbf{SV Agent}, and \textbf{EVAL Agent}, respectively. Agents invoked $\geq 2$ times are highlighted in gray.}
    \label{fig:1}
\end{figure*}

The proposed agentic system consists of two modules: planning and execution, as illustrated in Fig. \ref{fig:1}. To accommodate dynamic scene understanding tasks across different temporal scales, we categorize input scenes into three types based on their temporal length. Instantaneous scenes refer to single-frame image modalities at a specific time point, primarily used for static scene perception and semantic reasoning. For video modalities, drawing inspiration from computer vision, Video Question Answering (QA), and other related fields \cite{kim2023semi, yu2019activitynet}, we adopt 60 seconds as the temporal boundary and divide videos into two categories: short videos ($<$ 60 seconds) and long videos ($\geq$ 60 seconds), addressing the differing requirements of rapid-response and complex scene understanding tasks, respectively.

\subsection{Planning Module}

The execution flow of the planning module is detailed in Algorithm~\ref{alg:planning}. This module is powered by a Planning Agent based on either LLaVA~\cite{liu2023visual} or GPT-4o~\cite{hurst2024gpt} and operates under predefined prompts and a Chain-of-Thought (CoT) template. Its objective is to transform a natural language instruction $Q \in \mathcal{L}$ into a structured task and dispatch it to downstream execution agents.

\begin{algorithm}[!htbp]
\SetAlgoLined
\caption{Planning Module}
\label{alg:planning}
\KwIn{Natural language instruction $Q \in \mathcal{L}$; video $V$}
\KwOut{Final query set $Q$, extracted data $D$, modality label $\text{type}(D)$}

\While{$\text{time}(Q) = \varnothing$}{
  $Q \gets \text{Refine}(Q)$\;
}

\If{$Q$ is semantically ambiguous}{
  $Q \gets \text{CoT}(Q)$\;
}

Extract time $t \gets \text{time}(Q)$\;
Retrieve data: $D \gets \text{FFmpeg}(V, t)$ or $D \gets \text{FFmpeg}(V, [t_1, t_2])$\;

\eIf{$D$ is a single frame}{
  $\text{type}(D) \gets \text{Image}$\;
}{
  \eIf{$\text{duration}(D) < 60$}{
    $\text{type}(D) \gets \text{Short}$\;
  }{
    $\text{type}(D) \gets \text{Long}$\;
  }
}

\Return $Q$, $D$, and $\text{type}(D)$\;
\end{algorithm}

If the query lacks explicit temporal information (i.e., $\text{time}(Q) = \varnothing$), the system iteratively invokes an interactive refinement module to update $Q$ until a valid temporal reference is provided. For semantically ambiguous queries, the system applies a CoT template to decompose $Q$ into more specific sub-questions (${q_1, q_2, \dots, q_n}$), which replace the original query. Once temporal information $t$ is extracted from $Q$, the system uses the FFmpeg tool to retrieve either an image frame or a video clip from the input video $V$: $\text{FFmpeg}(V, t) \rightarrow I$ or $\text{FFmpeg}(V, [t_1, t_2]) \rightarrow C$, resulting in data $D$. The modality label of $D$ is automatically determined based on duration as follows:
\[
\text{type}(D) =
\begin{cases}
\text{Image}, & \text{if } D \text{ is a single frame} \\
\text{Short}, & \text{if } \text{duration}(D) < 60s \\
\text{Long}, & \text{if } \text{duration}(D) \geq 60s
\end{cases}
\]
The modality label $\text{type}(D)$, along with query $Q$ and data $D$, is forwarded to the execution module.

\subsection{Execution Module}

Upon completion of modality classification, the system dispatches the task to the corresponding execution agent. LLaVA or GPT-4o models are integrated into each agent to perform perception and reasoning.

\begin{algorithm}[!htbp]
\SetAlgoLined
\caption{Execution Module}
\label{alg:execution}
\KwIn{Data $D$ with modality label $\text{type}(D)$; user query $Q$; UAV parameters $f$, $v$, $\lambda$; candidate cluster set $\mathcal{K}=\{4,6,9,12,16,20,25\}$; prompt templates $\text{Prompt}_{\text{eval}}$, $\text{Prompt}_{\text{grid}}$.}
\KwOut{Answer $A$}
\Switch{\text{type}(D)}{
  \Case{\text{Image}}{
    $A \gets \text{AirVista}(I, Q)$\;
  }
  \Case{\text{Short}}{
    Extract keyframes: $F \gets \text{OpenCV}(C)$ with $F = \{f_1,f_2,\dots,f_6\}$\;
    Construct grid: $G \gets \text{Grid}(F)$\;
    $A \gets \text{Answer}(G, Q, \text{Prompt}_{\text{grid}})$\;
  }
  \Case{\text{Long}}{
    Compute sampling step: $s \gets \lfloor f \cdot \lambda / v \rfloor$\;
    Sample candidate frames: $F_{\text{cand}} \gets \{ f^{(i)} \mid f^{(i)} = C[i\cdot s] \}$\;
    Extract features: $Z \gets \{ \mathbf{z}_i \mid \mathbf{z}_i = \text{Feat}(f^{(i)}) \}$;
    \ForEach{$k \in \mathcal{K}$}{
      Compute clusters: $\mathcal{C}^{(k)} \gets \text{KMeans}(Z, k)$\;
      Evaluate clustering: $R_k \gets \text{Evaluate}(\mathcal{C}^{(k)}, \mathcal{M})$;
    }
    Select optimal cluster number: $K^* \gets \text{LLMSelect}(\{R_k\}_{k\in\mathcal{K}}, \text{Prompt}_{\text{eval}})$\;
    \ForEach{$i = 1$ to $K^*$}{
      $f^*_i \gets \arg\min_{f \in \mathcal{C}_i^{(K^*)}} \text{Time}(f)$\;
    }
    $F^* \gets \{f^*_1,f^*_2,\dots,f^*_{K^*}\}$\;
    Construct grid: $G \gets \text{Grid}(F^*)$\;
    $A \gets \text{Answer}(G, Q, \text{Prompt}_{\text{grid}})$\;
  }
}
\Return $A$\;
\end{algorithm}

\textbf{Instantaneous Image Tasks}: The Image Agent receives the image $I$ and user query $Q$ and invokes the AirVista \cite{lin2024airvista} tool, a UAV-specific multimodal question-answering model, to generate the answer $A$: $\text{AirVista}(I, Q) \rightarrow A$.

\textbf{Short Video Tasks}: The Short Video Agent first extracts 6 evenly spaced keyframes from short video $C$ using OpenCV, forming the set $F = \{f_1, f_2, \dots, f_6\}$, which are arranged into a $3 \times 2$ temporal grid image $G = \text{Grid}(F)$. Guided by the grid prompt $\text{Prompt}_{\text{grid}}$, the agent performs self-reasoning over $G$ and $Q$ to produce an answer: $\text{Answer}(G, Q, \text{Prompt}_{\text{grid}}) \rightarrow A$ \cite{kim2024image}. This strategy significantly reduces computational overhead while preserving temporal context.

\textbf{Long Video Tasks}: For videos longer than 60 seconds with complex scene dynamics, fixed-frame sampling is insufficient to capture semantic diversity. We propose an adaptive keyframe extraction strategy combining motion-aware sampling, clustering analysis, and model-guided selection.

The Long Video Agent first computes a sampling step $s = \lfloor f \cdot \lambda / v \rfloor$, where $f$ is the frame rate, $v$ is the average UAV velocity (m/s), and $\lambda$ is the desired semantic resolution (m/frame). This ensures that the UAV moves at least $\lambda$ meters between sampled frames, balancing coverage and efficiency \cite{rizzoli2023syndrone}. In our setup, $f = 25$, $v = 5$, and $\lambda = 5$. Using this, the agent samples candidate frames from video $C$: $F_{\text{cand}} = \{ f^{(i)} \mid f^{(i)} = C[i \cdot s] \}$.

Next, the agent applies a visual feature extractor $\text{Feat}(\cdot)$ (CLIP ViT-B/16 \cite{radford2021learning}) to compute high-dimensional semantic embeddings $Z = \{ \mathbf{z}_i \mid \mathbf{z}_i = \text{Feat}(f^{(i)}) \}$. To adaptively determine the number of keyframes, the agent performs $k$-means clustering for $k \in \mathcal{K} = \{4, 6, 9, 12, 16, 20, 25\}$: $\mathcal{C}^{(k)} = \text{KMeans}(Z, k)$. These values are selected to ensure that the keyframes can be arranged into near-square image grids (e.g., $3\times3$, $4\times4$, $5\times5$). As shown in the experiments of IG-VLM \cite{kim2024image}, square or near-square layouts better preserve visual details and temporal coherence, leading to improved performance in visual-language reasoning tasks compared to wide or tall grid arrangements. Each clustering result is evaluated using multiple metrics $\mathcal{M}$ (including Silhouette Score, Elbow Method, Calinski-Harabasz, and Davies-Bouldin indices): $R_k = \text{Evaluate}(\mathcal{C}^{(k)}, \mathcal{M})$.

These evaluation scores are passed to a GPT-4o-based evaluation agent, which analyzes and aggregates the metrics using a prompt template $\text{Prompt}_{\text{eval}}$, selecting the optimal number of clusters via contextual reasoning: $K^* = \text{LLMSelect}(\{R_k\}_{k \in \mathcal{K}}, \text{Prompt}_{\text{eval}})$. The Long Video Agent then selects the earliest-timestamped frame from each cluster $\mathcal{C}_i^{(K^*)}$ to form the final keyframe set: $F^* = \{ f^*_i \mid f^*_i = \arg\min_{f \in \mathcal{C}_i^{(K^*)}} \text{Time}(f) \}$, and constructs a near-square grid image $G = \text{Grid}(F^*)$.

Finally, with the grid prompt $\text{Prompt}_{\text{grid}}$, the agent performs over the grid and question: $\text{Answer}(G, Q, \text{Prompt}_{\text{grid}}) \rightarrow A$. The keyframes are capped at 25, which mitigates semantic degradation due to small object scales in aerial imagery and balances recognition accuracy with visual completeness.


\section{Experiments}

To validate the effectiveness of the proposed agentic system, we design experiments targeting both short and long video scenarios, with a focus on evaluating the system’s semantic understanding and question-answering capabilities under UAV dynamic perspectives. The experiments are constructed based on the ERA\cite{mou2020era}, CapERA \cite{bashmal2023capera}, and SynDrone \cite{rizzoli2023syndrone} datasets, and employ quantitative metrics to assess the accuracy and consistency of the generated answers. Considering that image-modality tasks are already supported by our previously proposed AirVista \cite{lin2024airvista}—which enables fine-grained semantic understanding and 3D spatial reasoning from UAV viewpoints—this section focuses on addressing the more challenging video-modality tasks.

\subsection{Short Video Scenario Experiments}

To evaluate the system's understanding capabilities in short video dynamic scenes, we constructed a content summarization QA task based on the CapERA dataset, named CapERA-QA. The original dataset contains five human-annotated captions per video; we randomly select one as the reference answer and manually construct a corresponding question in the unified form: \textit{``Please summarize the main content of this video in one concise sentence.''} This question format is designed to simulate real-world scenarios where UAVs are required to generate event-level summaries of video scenes.

We adopt the GPT-based semantic evaluation method proposed by Video-ChatGPT \cite{maaz2023video}. The evaluation results are shown in Table \ref{tab:1}. Count denotes the number of evaluated UAV video samples. Accuracy refers to the proportion of responses scored 4 or 5 by GPT-3.5, indicating semantically correct or fully correct answers. The score represents the average rating assigned by GPT-3.5, with a maximum of 5, reflecting overall semantic quality.

\begin{table}[!htbp]
\centering
\caption{GPT-based Evaluation Results on CapERA-QA Dataset}
\label{tab:1}
\begin{tabular*}{\linewidth}{@{\extracolsep{\fill}}lccc}
\toprule
Dataset    & Count & Accuracy & Score \\
\midrule
CapERA-QA  & 2864  & 0.756    & 3.703 \\
\bottomrule
\end{tabular*}
\end{table}

We further conducted manual comparisons on selected samples, and the results indicate that the model can accurately capture the main events and dynamic semantics in most aerial videos. Overall, the experiment demonstrates the effectiveness and practical potential of the proposed pipeline in UAV short-video question-answering tasks.

We further evaluated the readability of the system-generated answers to assess their clarity and comprehensibility in natural language expression. Specifically, we adopted several mainstream English readability metrics, including the Gunning Fog Index, Dale–Chall Readability Formula, Automated Readability Index, Coleman–Liau Index, and Linsear Write Formula. The statistical results are summarized in Table \ref{tab:2}, where the values in parentheses indicate the range of each metric across all samples, reflecting the variation in readability among different QA pairs.

\begin{table}[!htbp]
\centering
\caption{Readability Scores of System Answers on CapERA-QA}
\label{tab:2}
\begin{tabular*}{\linewidth}{@{\extracolsep{\fill}}lcc}
\toprule
Metric & Mean & Range \\
\midrule
Gunning Fog Index & 12.16 & 6.17--17.71 \\
Dale--Chall Score & 9.81 & 6.89--12.70 \\
Automated Readability Index & 11.26 & 4.30--18.05 \\
Coleman--Liau Index & 10.73 & 4.58--16.83 \\
Linsear Write Formula & 11.55 & 5.69--19.67 \\
\bottomrule
\end{tabular*}
\end{table}

\begin{figure}[!htbp]
    \centering
    \includegraphics[width=\columnwidth]{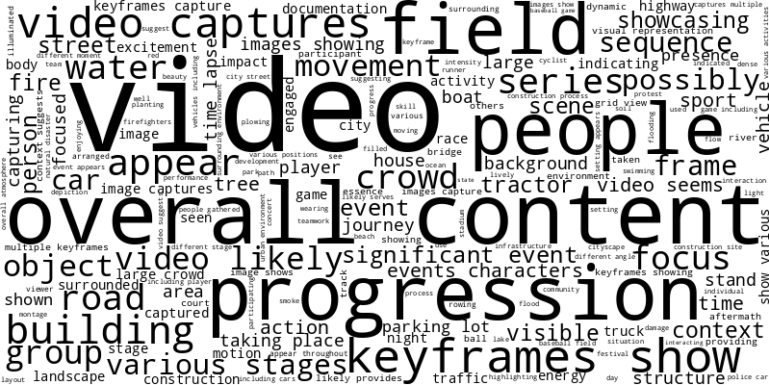}
    \caption{Word Cloud Visualization.}
    \label{fig:2}
\end{figure}

We further conducted a qualitative analysis of the generated answers by visualizing the high-frequency semantic content using a word cloud, as illustrated in Fig. \ref{fig:2}. This visualization highlights the dominant concepts captured by the system across diverse UAV video scenarios, providing intuitive insights into the model’s semantic focus.

To further evaluate the system’s capability for complex semantic understanding and reasoning in dynamic scenes, we construct an open-ended question-answering task named ERA-QA. This task is built upon the ERA dataset, from which 200 aerial videos with clear semantics and strong questionability were manually selected. Following the construction methodology of ActivityNet-QA \cite{yu2019activitynet}, question-answer pairs were designed accordingly. The question types cover four categories: motion understanding, spatial relations, temporal relations, and free-form questions. The answer types span six categories: yes/no, color, object, location, number, and others. We adopt the same GPT-based automatic evaluation strategy. Additionally, to analyze the impact of different VLMs on QA performance, we compare the short video agents based on LLaVA-1.6-34B \cite{liu2024llavanext} and GPT-4o (2024-05-13) \cite{hurst2024gpt}. The results are presented in Table \ref{tab:3}.

\begin{table}[!htbp]
\centering
\caption{GPT-based Evaluation Results on ERA-QA Dataset}
\label{tab:3}
\begin{tabular*}{\linewidth}{@{\extracolsep{\fill}}l l c c c}
\toprule
\textbf{Dataset} & \textbf{Backbone} & \textbf{Count} & \textbf{Accuracy} & \textbf{Score} \\
\midrule
\multirow{2}{*}{ERA-QA} 
& LLaVA-1.6-34B          & \multirow{2}{*}{200} & 0.665 & 3.715 \\
& GPT-4o (2024-05-13)    &                      & 0.530 & 3.140 \\
\bottomrule
\end{tabular*}
\end{table}

\subsection{Long Video Scenario Experiments}

The pipelines for short and long videos are largely identical, except for the adaptive keyframe extraction in the latter. The effectiveness of this pipeline in dynamic scene understanding has already been demonstrated in short video experiments. Therefore, the focus of the long-video experiments is to evaluate the practical effectiveness of the adaptive keyframe extraction strategy. We construct a long-video QA task based on the SynDrone dataset, which provides frame-level annotations. Specifically, we select 8 representative scenes (Town01–Town07 and Town10HD) and concatenate all frames in temporal order to form complete videos, using a fixed aerial viewpoint at an altitude of 20 meters. For each video, we manually design one open-ended question, following the QA style of ActivityNet-QA, to assess the system’s comprehensive QA capability in long-duration, multi-event dynamic scenarios.

\begin{figure}[htbp]
  \centering

  \begin{minipage}[b]{0.48\linewidth}
    \centering
    \includegraphics[width=\linewidth]{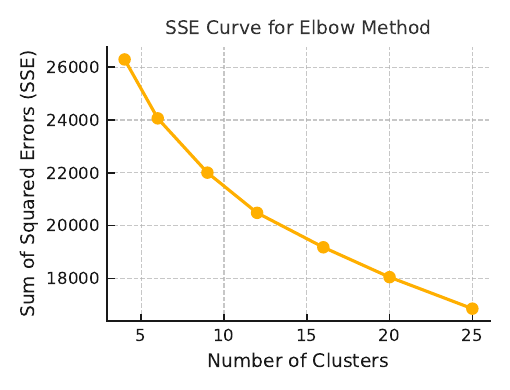}
    \vspace{-2em}
    
    {\scriptsize (a)}
  \end{minipage}
  \hfill
  \begin{minipage}[b]{0.48\linewidth}
    \centering
    \includegraphics[width=\linewidth]{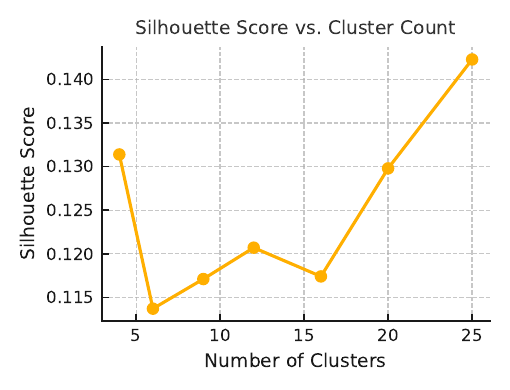}
    \vspace{-2em}

    {\scriptsize (b)}
  \end{minipage}

  \vspace{0.5em}

  \begin{minipage}[b]{0.48\linewidth}
    \centering
    \includegraphics[width=\linewidth]{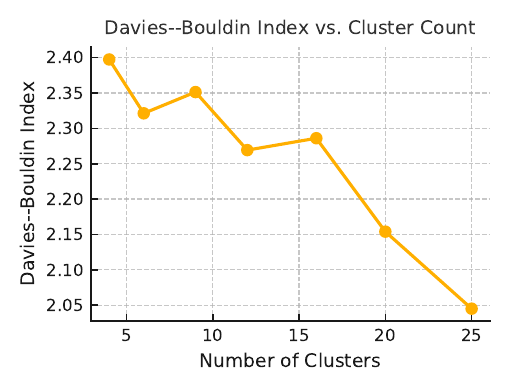}
    \vspace{-2em}

    {\scriptsize (c)}
  \end{minipage}
  \hfill
  \begin{minipage}[b]{0.48\linewidth}
    \centering
    \includegraphics[width=\linewidth]{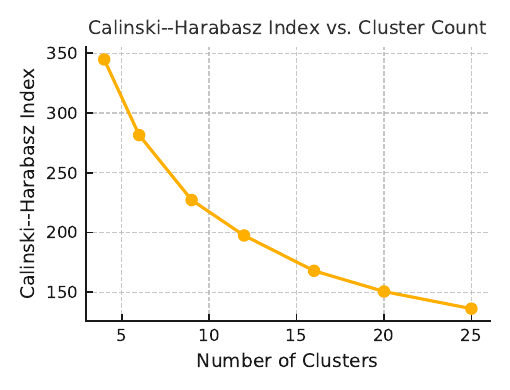}
    \vspace{-2em}

    {\scriptsize (d)}
  \end{minipage}

  \caption{Clustering evaluation results on Town01: (a) Sum of Squared Errors (SSE) curve for Elbow Method, (b) Silhouette Score, (c) Davies--Bouldin Index, and (d) Calinski--Harabasz Index.}
  \label{fig:3}
\end{figure}

To verify the clustering strategy, we conducted a visual analysis of the clustering evaluation results, which also served as a reference for the evaluation agent in selecting the optimal number of clusters $K$. The clustering evaluation metrics include the Silhouette Score, Elbow Method, Calinski–Harabasz index, and Davies–Bouldin index. Due to space limitations, Fig. \ref{fig:3} illustrates the metric curves for the Town01 scenario, where the final selected cluster number is 25, corresponding to a $5 \times 5$ image grid fed into the model. The statistics of the selected cluster numbers for the other scenarios are summarized in Table \ref{tab:4}.

\begin{table}[!htbp]
\centering
\caption{Clustering Results ($K^*$) for Each Long-Video Scene}
\label{tab:4}
\begin{tabular*}{\linewidth}{@{\extracolsep{\fill}}lcccc}
\toprule
\textbf{Scene} & \makecell{\textbf{Town01}\\\textbf{Town02}} & \makecell{\textbf{Town03}\\\textbf{Town04}} & \makecell{\textbf{Town05}\\\textbf{Town06}} & \makecell{\textbf{Town07}\\\textbf{Town10HD}} \\
\midrule
$K^*$ & 25 / 25 & 25 / 16 & 4 / 4 & 4 / 25 \\
\bottomrule
\end{tabular*}
\end{table}

As shown in Table \ref{tab:4}, our method adaptively selects significantly different numbers of keyframes based on scene complexity. For instance, in large-scale urban scenarios such as Town01, Town02, Town03, and Town10HD, where the video content is diverse and rapidly changing, the system tends to select a maximum of 25 frames to comprehensively capture key information. In contrast, for simpler and more static scenes such as Town05, Town06, and Town07, only 4 frames are sufficient to summarize the main content. In comparison, fixed-frame sampling strategies (e.g., uniformly selecting 6 frames) lack the flexibility to adjust to varying scene dynamics. For complex scenes like Town01, 6 frames may be insufficient to cover all critical events; for simple scenes, such a fixed number could lead to redundant frames and wasted computational resources. In contrast, our adaptive keyframe selection strategy dynamically scales the number of frames based on actual scene demands, demonstrating superior flexibility and efficiency.

To further validate the effectiveness of the adaptive keyframe selection strategy in long video QA tasks, we conducted a comparative experiment against a fixed-frame baseline. The baseline adopts the same strategy used in short video tasks, namely, uniformly sampling 6 frames. Both methods were evaluated on the same set of test videos. The statistical results are shown in Table 
 \ref{tab:5}.

\begin{table}[!htbp]
\centering
\caption{Comparison between fixed-frame and adaptive keyframe extraction on long video QA}
\label{tab:5}
\begin{tabular*}{\linewidth}{@{\extracolsep{\fill}}l c c c}
\toprule
\textbf{Method} & \textbf{Count} & \textbf{Accuracy} & \textbf{Score} \\
\midrule
Fixed 6-frame Sampling & \multirow{2}{*}{8} & 0.875 & 4.250 \\
Adaptive Keyframe Extraction    &                    & 1.000 & 4.375 \\
\bottomrule
\end{tabular*}
\end{table}

The results demonstrate the effectiveness of the adaptive keyframe selection strategy. By successfully capturing the essential semantic content of long videos, the approach enables the agent to generate complete and accurate responses. This finding further highlights the importance of incorporating adaptive perception modules for long-duration, dynamic scene understanding in embodied UAV systems.

\section{Conclusion}

This paper presents AirVista-II, an end-to-end agentic system designed to enhance the general semantic understanding and reasoning capabilities of embodied UAVs in dynamic environments. For long-duration videos, we designed an adaptive keyframe extraction method that effectively improves the system’s perception and reasoning performance for complex dynamic content. Future work will focus on optimizing the pipeline to reduce computational overhead and enhance overall system robustness.

\bibliographystyle{IEEEtran}
\bibliography{egbib}

\begin{thebibliography}{10}
\providecommand{\url}[1]{#1}
\csname url@samestyle\endcsname
\providecommand{\newblock}{\relax}
\providecommand{\bibinfo}[2]{#2}
\providecommand{\BIBentrySTDinterwordspacing}{\spaceskip=0pt\relax}
\providecommand{\BIBentryALTinterwordstretchfactor}{4}
\providecommand{\BIBentryALTinterwordspacing}{\spaceskip=\fontdimen2\font plus
\BIBentryALTinterwordstretchfactor\fontdimen3\font minus \fontdimen4\font\relax}
\providecommand{\BIBforeignlanguage}[2]{{%
\expandafter\ifx\csname l@#1\endcsname\relax
\typeout{** WARNING: IEEEtran.bst: No hyphenation pattern has been}%
\typeout{** loaded for the language `#1'. Using the pattern for}%
\typeout{** the default language instead.}%
\else
\language=\csname l@#1\endcsname
\fi
#2}}
\providecommand{\BIBdecl}{\relax}
\BIBdecl

\bibitem{wu2024embodied}
Y.~Wu, P.~Zhang, M.~Gu, J.~Zheng, and X.~Bai, ``Embodied navigation with multi-modal information: A survey from tasks to methodology,'' \emph{Information Fusion}, p. 102532, 2024.

\bibitem{tian2025algorithmic}
Y.~Tian, Y.~Zhang, and F.-Y. Wang, \emph{Algorithmic Foundations of Large Models: Principles and Applications of Transformers}.\hskip 1em plus 0.5em minus 0.4em\relax Beijing: Tsinghua University Press, 2025.

\bibitem{zheng2024computational}
W.~Zheng and F.-Y. Wang, \emph{Computational Knowledge Vision: The First Footprints}.\hskip 1em plus 0.5em minus 0.4em\relax Elsevier, 2024.

\bibitem{tian2024logisticsvista}
Y.~Tian, F.~Lin, X.~Zhang, J.~Ge, Y.~Wang, X.~Dai, Y.~Lv, and F.-Y. Wang, ``Logisticsvista: 3d terminal delivery services with uavs, ugvs and usvs based on foundation models and scenarios engineering,'' in \emph{2024 IEEE International Conference on Service Operations and Logistics, and Informatics (SOLI)}.\hskip 1em plus 0.5em minus 0.4em\relax IEEE, 2024.

\bibitem{de2023socratic}
I.~de~Zarza, J.~de~Curto, and C.~T. Calafate, ``Socratic video understanding on unmanned aerial vehicles,'' \emph{Procedia Computer Science}, vol. 225, pp. 144--154, 2023.

\bibitem{de2023semantic}
J.~De~Curt{\`o}, I.~De~Zarza, and C.~T. Calafate, ``Semantic scene understanding with large language models on unmanned aerial vehicles,'' \emph{Drones}, vol.~7, no.~2, p. 114, 2023.

\bibitem{zhao2025cityeqa}
\BIBentryALTinterwordspacing
Y.~Zhao, K.~Xu, Z.~Zhu, Y.~Hu, Z.~Zheng, Y.~Chen, Y.~Ji, C.~Gao, Y.~Li, and J.~Huang, ``Cityeqa: A hierarchical llm agent on embodied question answering benchmark in city space,'' 2025. [Online]. Available: \url{https://arxiv.org/abs/2502.12532}
\BIBentrySTDinterwordspacing

\bibitem{shavit2023practices}
Y.~Shavit, S.~Agarwal, M.~Brundage, S.~Adler, C.~O’Keefe, R.~Campbell, T.~Lee, P.~Mishkin, T.~Eloundou, A.~Hickey \emph{et~al.}, ``Practices for governing agentic ai systems,'' \emph{Research Paper, OpenAI}, 2023.

\bibitem{tian2025uavs}
Y.~Tian, F.~Lin, Y.~Li, T.~Zhang, Q.~Zhang, X.~Fu, J.~Huang, X.~Dai, Y.~Wang, C.~Tian, B.~Li, Y.~Lv, L.~Kov{\'a}cs, and F.-Y. Wang, ``Uavs meet llms: Overviews and perspectives towards agentic low-altitude mobility,'' \emph{Information Fusion}, vol. 122, p. 103158, 2025.

\bibitem{lin2023video}
\BIBentryALTinterwordspacing
B.~Lin, Y.~Ye, B.~Zhu, J.~Cui, M.~Ning, P.~Jin, and L.~Yuan, ``Video-llava: Learning united visual representation by alignment before projection,'' 2024. [Online]. Available: \url{https://arxiv.org/abs/2311.10122}
\BIBentrySTDinterwordspacing

\bibitem{maaz2023video}
\BIBentryALTinterwordspacing
M.~Maaz, H.~Rasheed, S.~Khan, and F.~S. Khan, ``Video-chatgpt: Towards detailed video understanding via large vision and language models,'' 2024. [Online]. Available: \url{https://arxiv.org/abs/2306.05424}
\BIBentrySTDinterwordspacing

\bibitem{kim2024image}
W.~Kim, C.~Choi, W.~Lee, and W.~Rhee, ``An image grid can be worth a video: Zero-shot video question answering using a vlm,'' \emph{IEEE Access}, 2024.

\bibitem{wang2024videotree}
\BIBentryALTinterwordspacing
Z.~Wang, S.~Yu, E.~Stengel-Eskin, J.~Yoon, F.~Cheng, G.~Bertasius, and M.~Bansal, ``Videotree: Adaptive tree-based video representation for llm reasoning on long videos,'' 2025. [Online]. Available: \url{https://arxiv.org/abs/2405.19209}
\BIBentrySTDinterwordspacing

\bibitem{park2024too}
\BIBentryALTinterwordspacing
J.~Park, K.~Ranasinghe, K.~Kahatapitiya, W.~Ryu, D.~Kim, and M.~S. Ryoo, ``Too many frames, not all useful: Efficient strategies for long-form video qa,'' 2025. [Online]. Available: \url{https://arxiv.org/abs/2406.09396}
\BIBentrySTDinterwordspacing

\bibitem{yao2024aeroverse}
\BIBentryALTinterwordspacing
F.~Yao, Y.~Yue, Y.~Liu, X.~Sun, and K.~Fu, ``Aeroverse: Uav-agent benchmark suite for simulating, pre-training, finetuning, and evaluating aerospace embodied world models,'' 2024. [Online]. Available: \url{https://arxiv.org/abs/2408.15511}
\BIBentrySTDinterwordspacing

\bibitem{yuan2024patrol}
Z.~Yuan, F.~Xie, and T.~Ji, ``Patrol agent: An autonomous uav framework for urban patrol using on board vision language model and on cloud large language model,'' in \emph{2024 6th International Conference on Robotics and Computer Vision (ICRCV)}.\hskip 1em plus 0.5em minus 0.4em\relax IEEE, 2024, pp. 237--242.

\bibitem{lin2024airvista}
F.~Lin, Y.~Tian, Y.~Wang, T.~Zhang, X.~Zhang, and F.-Y. Wang, ``Airvista: Empowering uavs with 3d spatial reasoning abilities through a multimodal large language model agent,'' in \emph{2024 IEEE 27th International Conference on Intelligent Transportation Systems (ITSC)}.\hskip 1em plus 0.5em minus 0.4em\relax IEEE, 2024, pp. 476--481.

\bibitem{mou2020era}
L.~Mou, Y.~Hua, P.~Jin, and X.~X. Zhu, ``Era: A data set and deep learning benchmark for event recognition in aerial videos [software and data sets],'' \emph{IEEE Geoscience and Remote Sensing Magazine}, vol.~8, no.~4, pp. 125--133, 2020.

\bibitem{bashmal2023capera}
L.~Bashmal, Y.~Bazi, M.~M. Al~Rahhal, M.~Zuair, and F.~Melgani, ``Capera: Captioning events in aerial videos,'' \emph{Remote Sensing}, vol.~15, no.~8, p. 2139, 2023.

\bibitem{rizzoli2023syndrone}
G.~Rizzoli, F.~Barbato, M.~Caligiuri, and P.~Zanuttigh, ``Syndrone-multi-modal uav dataset for urban scenarios,'' in \emph{Proceedings of the IEEE/CVF International Conference on Computer Vision}, 2023, pp. 2210--2220.

\bibitem{kim2023semi}
\BIBentryALTinterwordspacing
S.~Kim, J.-H. Kim, J.~Lee, and M.~Seo, ``Semi-parametric video-grounded text generation,'' 2023. [Online]. Available: \url{https://arxiv.org/abs/2301.11507}
\BIBentrySTDinterwordspacing

\bibitem{yu2019activitynet}
Z.~Yu, D.~Xu, J.~Yu, T.~Yu, Z.~Zhao, Y.~Zhuang, and D.~Tao, ``Activitynet-qa: A dataset for understanding complex web videos via question answering,'' in \emph{Proceedings of the AAAI Conference on Artificial Intelligence}, vol.~33, no.~01, 2019, pp. 9127--9134.

\bibitem{liu2023visual}
H.~Liu, C.~Li, Q.~Wu, and Y.~J. Lee, ``Visual instruction tuning,'' \emph{Advances in neural information processing systems}, vol.~36, pp. 34\,892--34\,916, 2023.

\bibitem{hurst2024gpt}
\BIBentryALTinterwordspacing
A.~Hurst, A.~Lerer, A.~P. Goucher, A.~Perelman, A.~Ramesh, A.~Clark, A.~Ostrow, A.~Welihinda, A.~Hayes, A.~Radford \emph{et~al.}, ``Gpt-4o system card,'' 2024. [Online]. Available: \url{https://arxiv.org/abs/2410.21276}
\BIBentrySTDinterwordspacing

\bibitem{radford2021learning}
A.~Radford, J.~W. Kim, C.~Hallacy, A.~Ramesh, G.~Goh, S.~Agarwal, G.~Sastry, A.~Askell, P.~Mishkin, J.~Clark \emph{et~al.}, ``Learning transferable visual models from natural language supervision,'' in \emph{International conference on machine learning}.\hskip 1em plus 0.5em minus 0.4em\relax PmLR, 2021, pp. 8748--8763.

\bibitem{liu2024llavanext}
\BIBentryALTinterwordspacing
H.~Liu, C.~Li, Y.~Li, B.~Li, Y.~Zhang, S.~Shen, and Y.~J. Lee, ``Llava-next: Improved reasoning, ocr, and world knowledge,'' January 2024. [Online]. Available: \url{https://llava-vl.github.io/blog/2024-01-30-llava-next/}
\BIBentrySTDinterwordspacing

\end{thebibliography}

\end{document}